\documentclass[11pt]{article}

\usepackage[preprint]{acl}

\usepackage{times}
\usepackage{latexsym}
\usepackage[T1]{fontenc}
\usepackage{amsmath,amssymb}
\usepackage{amsthm}
\usepackage{stmaryrd}  
\usepackage{float}
\usepackage[table]{xcolor}

\usepackage{booktabs}
\usepackage{tabularx}

\usepackage[T1]{fontenc}

\usepackage{listings}          
\lstdefinestyle{dsl}{          
  basicstyle=\ttfamily\footnotesize,
  columns=fullflexible,
  keepspaces=true,
  frame=single,
  framesep=6pt,
  xleftmargin=4pt,
  aboveskip=8pt,
  belowskip=8pt,
}

\usepackage[utf8]{inputenc}

\usepackage{microtype}

\usepackage{inconsolata}

\usepackage{graphicx}

\title{Improving Natural-Language Combinatorial-Optimization Accuracy in Resource-Constrained Language Models via Formal Abstractions}

\author{Shrenil Shaun Sharma \\
  Independent Researcher\\
  San Francisco, CA, USA\\
  \texttt{shrenil19+research@gmail.com} \\\And
  Avi Sharma \\
  Department of Electrical \\
  Engineering and Computer Sciences \\
  University of California, Berkeley \\
  \texttt{avi\_sharma@berkeley.edu} \\}

\begin{document}
\maketitle

\begin{abstract}

Combinatorial scheduling poses a significant challenge for language models, requiring them to identify feasible solutions within exponentially large search spaces while satisfying complex constraints. This challenge is especially pronounced in resource-constrained settings, where larger language models are impractical and selection is limited to smaller models which often fail to preserve feasibility when scheduling directly from natural language. To address these limitations, we introduce SDDL, a neuro-symbolic framework that translates natural-language scheduling problems into compact, solver-aligned representations of tasks, resources, constraints, and objectives, while delegating low-level modeling and search to a deterministic compiler and external solver. On a 300-instance, multi-family subset of scheduling problems, SDDL improves independently verified feasibility for every resource-constrained model tested. The two strongest SDDL configurations reach 55.3\% and 28.3\%, up from direct-generation baselines of 23.7\% and 1.3\% and solver-code baselines of 21.7\% and 7.0\%, with a 0.0\% median optimality gap among feasible schedules. By expressing problem structure rather than generating solutions or solver code, SDDL enables smaller models to approach the strongest evaluated direct- and solver-code configurations, including substantially larger frontier models.

\end{abstract}
\section{Introduction}

Many requests posed to modern language models arrive in ordinary language, including problems whose solutions depend on an underlying mathematical structure. In such cases, constraints, objectives, and procedures may be specified only implicitly, requiring the model to infer the mathematical problem being described. For tasks that instantiate optimization problems, this inference is not merely semantic: the model must translate a verbal description into a latent search space of possible solutions and evaluate candidates against implicit feasibility and optimality criteria. This challenge is especially pronounced in resource-constrained settings, where model selection is limited to language models with substantially fewer parameters than larger available alternatives. Generating a solution requires parsing the task description, tracking interacting constraints, reasoning about objectives, and implicitly searching over alternatives within a single autoregressive pass. For smaller models, this process often yields fluent but infeasible solutions.

Neuro-symbolic decomposition may  mitigate this limitation by separating language understanding from downstream computation: the language model produces an executable or formal representation, and an external runtime or solver performs the corresponding computation \citep{gao2023pal,pan2023logiclm}. In optimization-specific systems, this formalization may include decision variables, constraints, and objectives \citep{ramamonjison2022nl4opt,ahmaditeshnizi2024optimus,shi2025constraintllm}. This approach, however, shifts the bottleneck to translation fidelity. A solver optimizes only the formalization it receives, so omitted constraints, misdefined variables, or distorted objectives directly undermine the resulting solution. Combinatorial optimization problems expose this weakness particularly well, where unlike small decision problems or clue-based logic puzzles, instances must represent objectives, resource capacities, temporal relations, precedence constraints, and disjunctive alternatives at scale. We find that direct solver-code generation by resource-constrained models often fails as a formalization strategy, producing programs that may be executable and solver-feasible yet unfaithful to the intended problem. To bridge this gap, we introduce \textbf{SDDL (Scheduling Domain Definition Language)}, a domain-specific language that narrows the translation target from open-ended solver code to a small set of solver-aligned scheduling primitives. Rather than requiring models to emit low-level solver code, SDDL provides scheduling-native abstractions for tasks, resources, constraints, alternatives, and objectives, where SDDL programs are deterministically compiled into a solver model for execution by an external solver. Across 13 models we compare direct and generic solver-code generation, and evaluate \textbf{SDDL} across multiple resource-constrained models. SDDL improves independently verified feasibility for every model tested with it, while reducing the median optimality gap among feasible schedules. Its strongest result ranks among the strongest configurations evaluated overall, while others improve to several times their baselines, demonstrating the effectiveness of SDDL relative to direct and generic solver-code generation, and enabling smaller models to match stronger configurations. Our contributions are:

\begin{enumerate}
\item \textbf{SDDL (Scheduling Domain Definition Language)} a domain-specific language for executable, objective-bearing combinatorial scheduling formulations that improves the fidelity of resource-constrained LLM formalization relative to direct solver-code generation.
\item An independently verified evaluation of \textbf{SDDL} against generation strategies matched in conceptual scope and solver target.
\item \textbf{An evaluation showing SDDL enables substantially smaller models to match} or closely approach the performance of the strongest direct- and solver-code configurations evaluated, including those using substantially larger frontier models.

\end{enumerate}

\section{Related Work}

\subsection{Structured Formal Reasoning}

Empirical evaluations such as PlanBench document substantial weaknesses in systematic, multi-step planning \citep{valmeekam2023planbench}, while \citet{kambhampati2024llms} argue that LLMs are better incorporated into frameworks that delegate planning to external modules. Performance on constrained-generation tasks also varies substantially across model scales, with smaller, resource-constrained models achieving lower constraint-satisfaction rates than larger models, including under zero-shot prompting \citep{yao2024collie}. This gap may be compounded by restrictive output-format requirements, which have been shown to degrade performance on reasoning-heavy tasks \citep{tam2024let}. One such formulation arises in combinatorial scheduling problems, where models must simultaneously recover problem semantics, respect representational conventions, and maintain constraint fidelity across interdependent decisions and temporal relations. Together, these findings suggest that preserving reasoning capacity during inference may depend on reducing what the model must formalize and compute, motivating approaches that delegate execution to external tools.

\subsection{Solver-Delegated Constraint Reasoning}

Neuro-symbolic frameworks leverage LLMs to parse unstructured text into executable representations, delegating computation to external runtimes to bypass internal arithmetic and logical errors \citep{gao2023pal,chen2023program}. Logic-LM extends this approach to symbolic inference, translating natural-language problems into formal logic and using solver feedback to iteratively repair invalid representations \citep{pan2023logiclm}. These results indicate that LLMs can benefit from constructing executable or symbolic representations while delegating deterministic computation and inference to external tools. NL4Opt \citep{ramamonjison2022nl4opt}, OptiMUS \citep{ahmaditeshnizi2024optimus}, and ConstraintLLM \citep{shi2025constraintllm} apply this paradigm to optimization and constraint programming at different levels of abstraction. However, OptiMUS and ConstraintLLM target highly expressive, general-purpose programming environments, where formalization can require verbose variable declarations, low-level solver API calls, and, often, explicit control flow on top of translating the underlying problem semantics; generality that may come at the cost of reliable generation.

\subsection{Solver-Free Generation and Decoding}

Alternative approaches evaluate LLMs as direct end-to-end combinatorial solvers \citep{jiang-etal-2026-reasoning-combinatorial}. In such end-to-end generation contexts, evaluation shows that solution quality and feasibility degrade as instance size and structural complexity increase \citep{jiang-etal-2026-reasoning-combinatorial}. Separately, structured-decoding approaches constrain the generation process itself: syntax-aware parsing \citep{yin2018tranx} and grammar-constrained decoding \citep{geng2023grammar} enforce structural validity at the token level. However, hard formatting constraints can degrade underlying reasoning performance \citep{tam2024let}, and a syntactically valid representation may still omit a critical constraint or transfer an incorrect value. SDDL therefore rejects malformed programs at parse time and verifies emitted schedules against the source instance, ensuring syntactic validity alone is not treated as evidence of correctness.

\subsection{LLMs for Combinatorial Scheduling}

While the paradigms detailed above address general constraint satisfaction, literature targeting scheduling-native problem structure remains narrow and fragmented. Existing scheduling applications mimic the solver-free paradigm through supervised fine-tuning for job-shop domains \citep{abgaryan2024llmsschedule}. Such approaches collapse problem interpretation and combinatorial search into a single generation step, making it difficult to attribute errors in an infeasible schedule to faulty constraint translation rather than search failure. Conversely, general-purpose CP systems relevant to scheduling \citep{michailidis2024constraint,shi2025constraintllm} inherit the heavy formalization overhead of general solver APIs and are not designed to exploit the recurring structural patterns (precedences, resource capacities, coverage requirements, penalized soft constraints) that span scheduling families. Closest to our setting, Logic.py \citep{kesseli2025logicpy} formalizes search-based problems through a  DSL for constraint solving and is evaluated primarily on logic-grid puzzles. Its evaluated system focuses on finding satisfying assignments, rather than optimizing objectives over feasible schedules.

\subsection{Natural-Language Scheduling Benchmarks}

Scheduling-specific natural-language benchmarks remain limited. Starjob \citep{abgaryan2025starjob} provides a large supervised corpus for end-to-end JSSP scheduling, but verbalizes its instances through fixed templates. R-ConstraintBench \citep{jain2025rconstraintbench} evaluates RCPSP feasibility under systematically varied constraints using similarly structured, field-like descriptions. NL $\Rightarrow$ Schedule \citep{liao-etal-2026-nl} offers fuller natural-language descriptions through semi-synthetic instances constructed from real-world materials across four domains. NLCO \citep{jiang-etal-2026-reasoning-combinatorial} covers a broader range of combinatorial optimization families outside scheduling and only provides minimally verbalized instances. As our evaluation emphasizes formalization, we prioritize using a benchmark with unambiguous descriptions and canonical source instances, enabling generated schedules to be verified directly against formal ground truth. SCHEDBench \citep{2026schedbench} provides the best guarantee for this requirement by construction, where each description is a controlled, constraint-preserving verbalization of a canonical source instance drawn from established scheduling literature; providing a definitive formal ground truth and best-known objective for verifying  generated schedules.


\section{Methods and DSL Creation}
SDDL's design is based on an observation that automated formalization tends to break not only in understanding the problem, but in the many decisions required to render that understanding effectively \citep{shi2025constraintllm}. SDDL removes the failure-prone modeling decisions otherwise left to the model by naming only recurring scheduling structures for the model to identify, while a deterministic compiler handles their downstream encoding. We articulate first principles underlying this stance, then develop the interface that realizes them and the compilation guarantees it provides.

\begin{figure}[t]
\centering\footnotesize
\fbox{\parbox{0.93\linewidth}{\raggedright
\textit{JSSP instance, 10 jobs $\times$ 5 machines}
\vspace{1pt}\hrule\vspace{2pt}

\textbf{Rules (excerpt).}~Steps within an item follow the given order;
a location processes at most one step at a time; non-preemptive.\\[2pt]

\textbf{Setup.}~10 scenes on 5 crew stations. Each scene is an ordered
sequence of steps; each step names a station and a duration.\\[2pt]

\textbf{Scene Franklin} has a 5-step sequence. Step 1 is at the
\textit{North Sound Stage} for 12 hours; Step 2 is at the
\textit{Second Unit Screening Room} for 94 hours\ldots {\itshape[+ 9 more scenes]} \\[2pt]
\vspace{2pt}\hrule\vspace{1pt}

\textbf{Response format:} \texttt{<ItemName> step k: start=<int>}, one per line.
}}
\caption{Example scheduling problem rendered in natural-language; model must recover the underlying structure and emit a start time per step.}
\label{fig:jssp-example}
\end{figure}

\subsection{Scheduling Problems}
Scheduling problems require allocating activities to specific time and resource assignments to produce a feasible schedule that satisfies a set of constraints and optimizes an objective. Although they vary widely in form, their constraint structures are drawn from a few recurring relations: precedence between activities, disjunctive resources that process one activity at a time, cumulative resources that admit concurrency within a capacity, and multiple possible execution modes for each activity. Different problem families combine these differently: job-shop scheduling (JSSP) chains operations by precedence over disjunctive machines, whereas resource-constrained project scheduling (RCPSP) replaces machines with cumulative resources, and its multi-mode extension adds mode selection under nonrenewable budgets; all NP-hard, with makespan as the standard objective. Instances are constraint-dense (ex. Fig. \ref{fig:jssp-example}), so a single omitted or misread relation silently changes the feasible region, making scheduling a natural and demanding target for studying faithful formalization of natural-language problem statements.

\subsection{Design Principles}
We optimize the language for three properties: robustness, ensuring the modeling surface admits few malformed programs; concision, allowing constraints to be expressed without solver-specific machinery; and sufficient expressiveness for the intended problem families. We further require bounded expressiveness: rather than exposing a general-purpose language in which the model may introduce arbitrary variables or predicates, SDDL fixes a vocabulary of recurring scheduling constructs, precedence, disjunctive and cumulative resources, alternative execution modes, alongside supported objectives. This helps reduce opportunities for formalization errors, where otherwise, each degree of freedom exposed to the model creates another opportunity for errors (variable declarations, domain bounds, etc.) none of which specified by the original problem. By fixing the vocabulary, SDDL transfers these decisions to the compiler, recasting the model's role from synthesizing constraints to recognizing pre-constrained scheduling constructs. These properties stem from a deliberate design decision, as fixed grammatical primitives yield a closed surface language, parse-time rejection, and verified compilation. This also improves solver performance, as CP-SAT’s global constraints such as \texttt{NoOverlap} and \texttt{Cumulative}, propagate more strongly than equivalent Boolean decompositions. A free-form encoding may represent a cumulative resource correctly yet bury it in logic that the solver cannot recognize. Because SDDL names these constructs directly, the compiler can consistently emit the strongest global encoding. This bounded vocabulary both prevents formalization errors and preserves propagation strength. The remaining principles then cover a closed surface that rejects malformed programs at parse time, hides solver implementation details, and provides complete coverage without leaving constructs implicit.

\subsection{Primitives}

\begin{figure}
\centering
\footnotesize
\[
\begin{array}{r@{\;}c@{\;}l}
\textit{program}   & ::= & \textit{statement}^{*} \\[1pt]
\textit{statement} & ::= & \texttt{task}(\textit{id}, \textit{props}) \\
   & \mid & \texttt{resource}(\textit{id}, \textit{props}) \\
   & \mid & \texttt{before}(\textit{id}, \textit{id}) \\
   & \mid & \texttt{no\_overlap}(\textit{id}) \\
   & \mid & \texttt{conflict}(\textit{ids}, \texttt{group}{=}\textit{s}) \\
   & \mid & \texttt{not\_at}(\textit{id}, \textit{props}) \\
   & \mid & \texttt{penalize}(\textit{m}, \texttt{weight}{=}\textit{w}, \textit{props}) \\[1pt]
\textit{props}     & ::= & \epsilon \;\mid\; \textit{prop}\,(\texttt{,}\,\textit{prop})^{*} \\
\textit{ids}       & ::= & \textit{id}\,(\texttt{,}\,\textit{id})^{*} \\
\textit{prop}      & ::= & \textit{key}\,\texttt{=}\,\textit{value} \\
\textit{value}     & ::= & \textit{scalar} \mid\ \texttt{[}\,\textit{value},\dots\,\texttt{]}
                            \mid\ \texttt{\{}\,\textit{value}\,\texttt{:}\,\textit{value},\dots\,\texttt{\}} \\
\end{array}
\]
\caption{The SDDL grammar: ${}^{*}$ means ``zero or more,'' $\epsilon$ the empty string, and $\mid$ separates alternatives.
The atoms are \textit{id}, a quoted identifier; \textit{key}, a property name;
\textit{m}, a penalty measure; \textit{w}, an integer weight; \textit{s}, a group
name; and \textit{scalar}, a number or string.}
\label{fig:grammar}
\end{figure}

SDDL defines programs which consist of flat sequences of primitive scheduling constructs, with neither control flow, nor an expression language beyond literal values. Literals include "lists" and "maps", so values such as the \texttt{modes} list and \texttt{demands} and \texttt{consumes} maps are specified directly rather than constructed through code. The complete grammar is shown in Fig. \ref{fig:grammar} above.

Two of the seven primitives, \texttt{task} and \texttt{resource}, declare problem objects; four express hard constraints while \texttt{penalize} defines soft objectives. Constraint hardness is encoded structurally through the choice of primitive: a clause is hard unless it appears as a penalty, so the model never signals status through weights or phrasing. We selected the seven primitives to maximize coverage while minimizing the language surface, adding one only when existing constructs could not represent the required scheduling pattern. Two design choices illustrate this principle: precedence, which is expressed by the binary \texttt{before} primitive, capturing the pairwise finish-to-start ordering commonly used in these problems, avoiding a general temporal operator that the compiler could not otherwise translate uniformly. Second is shared-membership constraints, such as those between courses in the same curriculum, which are expressed using \texttt{conflict} with an optional \texttt{group} label. The compiler uses this label both to prohibit concurrent assignments, and to identify group-level objective terms. The remaining hard-constraint primitive, \texttt{not\_at}, forbids a task from executing during specified absolute times.

\begin{lstlisting}[style=dsl]
task(id, **props)          resource(id, **props)
before(a, b)               no_overlap(r)
conflict(t1, ..., group=)  not_at(id, **when)
penalize(measure, weight=w, **params)
\end{lstlisting}

\noindent{As an illustration, the clause \emph{``operation A runs on machine $m_0$ ...''} formalizes to the below primitives,}
\begin{lstlisting}[style=dsl]
task("a", machine="m0", duration=3)
task("b", machine="m1", duration=2)
before("a", "b")
\end{lstlisting}
whose lowering schedules the two tasks as intervals on their respective machines, then adds the single constraint $\mathit{start}_b \ge \mathit{end}_a$; with the model stating the operations and ordering, while every solver variable is introduced by the compiler.

\subsection{Property Polymorphism}

To cover varied task structures with a single declaration form, SDDL assigns \texttt{task} a property-dependent representation, with each task’s form inferred directly from its properties by the compiler. A task is \emph{continuous} if it specifies a duration or modes and no counts. Continuous tasks lower to start, end, and interval variables. This is especially useful for multi-mode activities, where the model supplies a list of duration-and-resource profiles while the compiler introduces the selection variables, optional intervals, and exactly-one constraint needed to choose a single mode. A second structure, discrete tasks declared by count, lowers to day-period-room variables for timetabling families; while specified and compiler-verified, it currently lies outside our evaluated scope. Because well-formed tasks must satisfy exactly one rule, archetype assignment is therefore deterministic, and the model therefore transcribes properties from the problem statement rather than choosing an internal representation. Property polymorphism therefore keeps the language compact while moving error-prone encoding decisions into the compiler.


\subsection{Resource Semantics}

Resource handling is where the language most clearly justifies its design as it makes explicit a distinction that natural-language descriptions often obscure but solvers must encode differently. A machine serving one job at a time is disjunctive, so assigned intervals cannot overlap. A shared worker pool is cumulative, so tasks may overlap provided total demand does not exceed capacity. Leaving this distinction implicit is a common source of modeling errors. The DSL resolves this ambiguity at declaration time. Renewable resources define a capacity referenced through a per-task \texttt{demands} map; their units are occupied during execution and released at completion, so they compile to cumulative constraints. Nonrenewable resources define a total and use a \texttt{consumes} map; their units are permanently expended, and compile to a project-wide linear budget. Disjunctive resources use \texttt{no\_overlap}. The explicit structure determines the encoding: \texttt{demands} indicates cumulative capacity, \texttt{consumes} indicates exhaustible supply, and \texttt{no\_overlap} indicates one-at-a-time use. The model need only classify the resource as capacity-limited, exhaustible, or strictly disjunctive, and the compiler then generates the corresponding constraint and prevents encoding errors. All 3 resource classes are exercised in our evaluation.

\subsection{Objectives}
Where hard constraints determine feasibility, \texttt{penalize} optimizes quality within the feasible region. Each penalty specifies a measure and weight; the compiler scales and sums the measures into a single minimized objective. Measures are selected from a fixed set: completion time, over-capacity, day spread, isolation, and room instability; rather than defined as free-form cost functions. This restriction provides the same safeguards as the closed constraint language. Arbitrary objectives can be syntactically valid yet semantically incorrect because of counting errors or optimization in the wrong direction. Named measures are implemented once by the compiler and reused consistently. The model therefore expresses only the intended preference and its weight, while the compiler realizes the objective correctly.

\subsection{Compilation}

Compilation is deterministic and requires no further model input. Programs are parsed through the language’s abstract-syntax machinery accepting only literals, so generated text is never executed, programs do not introduce side effects, and malformed inputs are rejected. Correctness is defined by an abstract, solver-independent semantics specifying schedules and objective values. A schedule assigns start and end times to continuous tasks, including selected durations for multi-mode tasks, and day-period-room tuples to discrete meetings. Each statement defines a feasibility condition, while \texttt{penalize} defines a weighted objective. For example, \texttt{before(a,b)} requires (a) to end before (b) starts, \texttt{no\_overlap} forbids simultaneous resource use, and a renewable \texttt{resource} of capacity (k) limits total active demand to (k). The compiler lowers each statement into solver variables, constraints, and an objective. Correctness requires the compiled model to preserve both feasibility and objective values. Direct primitives map immediately to solver constraints; cumulative resources, mode selection, and penalty measures involving reified constraints require additional equivalence arguments, provided in Appendix \ref{app:correctness}. For continuous tasks, the inferred horizon is ($H=T+\sum_i\max_{m\in M_i} d_{im}$), where (T) is one plus the latest forbidden time and (M\_i) is task (i)’s set of modes. Every feasible instance admits a serial schedule within this horizon; therefore, under the evaluated makespan objective, the horizon contains at least one optimal schedule.

\subsection{Guarantees and Robustness}
A closed vocabulary and deterministic compiler provide two guarantees normally unavailable when each instance is independently translated into solver code: every well-formed program compiles to a unique solver model without a new translation for each problem, and adequacy ensures that compilation preserves its meaning. Any feasible solution can be read as a horizon-bounded schedule satisfying the program, and every such schedule corresponds to a feasible solver solution with the same objective value. As the reference semantics and compiled model are constructed statement by statement, adequacy can be proven for each statement independently. For well-formed programs, adequacy removes compiler lowering as a source of semantic discrepancy, leaving the model-produced formalization as the remaining source of semantic error. The failure surface is therefore minimized, with every reported schedule tested directly against the source instance by an independent verifier. Extension beyond evaluated families requires no redesign of the language: discrete timetabling reuses the same declaration form through a second task archetype, and family-specific objectives are added by registering named measures in the existing compiler.

\section{Experimental Evaluation}
We use publicly released benchmark instances, evaluation harness, and verifiers to evaluate SDDL. We include previously reported DIRECT and SOLVER results from SCHEDBench, with newly generated results marked explicitly for contextual comparisons to characterize performance across model capabilities and generation modalities, while SDDL tests DSL-assisted combinatorial optimization in resource-constrained models. All conditions use fixed, instance-independent zero-shot prompts, greedy decoding (temperature 0), and no worked examples. Transient API failures are retried until generation completes, irrespective of solution quality. DIRECT includes only task and format instructions. SOLVER provides an OR-Tools CP-SAT guide covering equivalent scheduling concepts, conventions, and solver targets as SDDL; SDDL additionally includes its language specification due to its absence from pretraining corpora. SOLVER and SDDL therefore share conceptual scope but differ in formalization interface: open-ended solver code versus a closed language with deterministic compilation. Appendix \ref{model} gives the full inference settings, sandbox configuration, and pinned software and OR-Tools versions.


\subsection{SCHEDBench Evaluation Subset}
\label{sec:ex-setup}

We evaluate SDDL on the JSSP, single-mode RCPSP, and multi-mode RCPSP families of SCHEDBench, which dominate standard benchmarks and prior work on LLM scheduling. These families exercise the precedence, disjunctive and cumulative resources, nonrenewable budgets, multi-mode selection, and the objective channels of the DSL, and use a standardized makespan objective \texttt{penalize}, making optimality gap a clean measure of end-to-end feasibility; conflict, not\_at, and the discrete archetype are specified and compiler-verified but left unexercised. We exclude families scored through benchmark-specific weighted soft violations, whose instance specific penalty terms could be supported by registering additional named measures in the existing compiler.


\subsection{Evaluation Scoring and Metric Definitions}

Our primary metric is feasibility, where an emitted schedule must satisfy every hard constraint of the canonical source instance, as determined by an independent verifier. The pipeline renders a schedule as a structured listing of integer start times, checked directly against the source instance rather than the solver’s reported status. Violations are classified as precedence, machine-overlap, or resource-capacity errors. For feasible schedules, the verifier recomputes the objective from the schedule itself, preventing an incorrectly encoded objective from inflating performance. We then report the gap to the canonical optimum or best-known solution; where \emph{optimal} denotes a feasible schedule with a gap of at most ($10^{-9}$). Gaps are summarized by the median over feasible runs. Each run receives exactly 1 outcome: \emph{feasible}, \emph{infeasible}, \emph{no-solution}, or \emph{run-fail}. Failures before verification, including unparseable DSL output, transpilation errors, rendering failures, solver-reported infeasibility, and timeouts; are recorded as \emph{no-solution} or \emph{run-fail}. Feasibility therefore evaluates the complete path from natural-language input to verified schedule, not just the solver’s assessment of its own model. We report 95\% Wilson CIs for feasibility rates.  The DSL is parsed using a literal-only parser, never executed directly.

\subsection{Constraint Solving via SAT-Based Solvers}
Constraint programming (CP) represents a combinatorial problem using finite-domain decision variables and constraints over joint assignments. Search is interleaved with propagation, which reduces variable domains until a solution is found, infeasibility is proved, or an objective bound is certified. Modern SAT-based solvers implement this through lazy clause generation: propagators express their inferences as clauses for a conflict-driven SAT engine, combining CP propagation with SAT clause learning and linear relaxations for objective bounds. We target CP-SAT, the SAT-based constraint solver in Google OR-Tools. For scheduling, it provides interval variables linking a task’s start, duration, and end, with global constraints such as \texttt{NoOverlap} for disjunctive resources and \texttt{Cumulative} for shared-capacity resources. Their dedicated propagation methods, including overload checking, edge-finding, and energetic reasoning, prune more effectively than pairwise Boolean decompositions. CP-SAT also supports reified linear constraints and \texttt{AllDifferent} for assignment problems. Given an integer objective, CP-SAT proves optimality, returns the best solution found within the time limit, or proves infeasibility. Our compiler targets this interface directly, so SDDL primitives map to the solver constructs with the strongest relevant propagation.

\definecolor{sdqwen}{HTML}{0F62FE}
\definecolor{sddev}{HTML}{D9480F}
\begin{table*}[t]
\centering
\footnotesize
\setlength{\tabcolsep}{3.35pt}
\begin{tabular}{@{}l rrcr@{\hspace{3pt}}r @{\hspace{14pt}} rrcr @{\hspace{10pt}} r@{}}
\toprule
& \multicolumn{5}{c@{\hspace{14pt}}}{\textbf{Solver-Mediated Generation}} & \multicolumn{4}{c@{\hspace{10pt}}}{\textbf{Direct Generation}} & \\
\cmidrule(lr){2-6} \cmidrule(lr){7-10}
\textbf{Model} & $N$ & {\shortstack{\textbf{Feas.}\\(\%)}} & {\shortstack{\textbf{95\% CI}\\(\%)}} & {\shortstack{\textbf{R.\,fail}\\(\%)}} & {\shortstack{\textbf{Med.\ gap}\\(\%)}} & $N$ & {\shortstack{\textbf{Feas.}\\(\%)}} & {\shortstack{\textbf{95\% CI}\\(\%)}} & {\shortstack{\textbf{Med.\ gap}\\(\%)}} & {\shortstack{\textbf{$\Delta$\ Feas.}\\{}}} \\
\midrule
claude-opus-4-6 & 300 & 56.7$^{\ddagger}$ & {\scriptsize [51.0, 62.2]} & 21.0 & 0.0 & 300 & 4.7 & {\scriptsize [2.8, 7.7]} & 18.3 & \textbf{+52.0} \\
\rowcolor{black!8} \textbf{qwen3.5-27b $+$ SDDL}$^{\dagger}$ & 300 & \textcolor{sdqwen}{\textbf{55.3}}$^{\ddagger}$ & {\scriptsize [49.7, 60.9]} & 16.0 & 0.0 & 300 & \textcolor{sdqwen}{\textbf{23.7}} & {\scriptsize [19.2, 28.8]} & 395.8 & \textbf{+31.7} \\
gpt-5.5 (2026-04-23) & 300 & 53.3$^{\ddagger}$ & {\scriptsize [47.7, 58.9]} & 5.0 & 0.0 & 300 & 57.0 & {\scriptsize [51.3, 62.5]} & 36.7 & $-$3.7 \\
gpt-5.4 (2026-03-05) & 300 & 51.7 & {\scriptsize [46.0, 57.3]} & 18.3 & 0.0 & 300 & 0.3 & {\scriptsize [0.1, 1.9]} & 16.7 & \textbf{+51.3} \\
gpt-5.4-mini (2026-03-17) & 300 & 36.7 & {\scriptsize [31.4, 42.3]} & 32.3 & 7.5 & 300 & 0.7 & {\scriptsize [0.2, 2.4]} & 131.8 & \textbf{+36.0} \\
claude-sonnet-4-6 & 300 & 35.0$^{\ddagger}$ & {\scriptsize [29.8, 40.6]} & 24.3 & 0.0 & 300 & 8.3 & {\scriptsize [5.7, 12.0]} & 19.5 & +26.7 \\
qwen/qwen3.5-397b (2026-02-16) & 300 & 33.3 & {\scriptsize [28.2, 38.8]} & 12.3 & 0.0 & 300 & 19.3 & {\scriptsize [15.3, 24.2]} & 14.6 & +14.0 \\
\rowcolor{black!8} \textbf{devstral-small-2-24b $+$ SDDL}$^{\dagger}$ & 300 & \textcolor{sddev}{\textbf{28.3}}$^{\ddagger}$ & {\scriptsize [23.5, 33.7]} & 30.0 & 0.0 & 300 & \textcolor{sddev}{\textbf{1.3}}$^{\ddagger}$ & {\scriptsize [0.5, 3.4]} & 111.2 & +27.0 \\
qwen/qwen3.5-122b (2026-02-24) & 300 & 23.3 & {\scriptsize [18.9, 28.4]} & 56.0 & 0.6 & 300 & 7.0 & {\scriptsize [4.6, 10.5]} & 7.1 & +16.3 \\
\rowcolor{black!8} qwen/qwen3.5-27b (2026-02-24) & 300 & \textcolor{sdqwen}{\textbf{21.7}} & {\scriptsize [17.4, 26.7]} & 62.7 & 2.6 & 300 & \textcolor{sdqwen}{\textbf{23.7}} & {\scriptsize [19.2, 28.8]} & 395.8 & $-$2.0 \\
gemini-3.1-flash-lite & 300 & 17.0 & {\scriptsize [13.2, 21.7]} & 56.7 & 0.0 & 300 & 0.3 & {\scriptsize [0.1, 1.9]} & 107.1 & +16.7 \\
gemini-3-flash-preview & 300 & 12.0$^{\ddagger}$ & {\scriptsize [8.8, 16.2]} & 74.3 & 0.0 & 300 & 22.0 & {\scriptsize [17.7, 27.0]} & 24.5 & \emph{$-$10.0} \\
\rowcolor{black!8} devstral-small-2-24b (2025-12-09) & 300 & \textcolor{sddev}{\textbf{7.0}}$^{\ddagger}$ & {\scriptsize [4.6, 10.5]} & 66.7 & 0.0 & 300 & \textcolor{sddev}{\textbf{1.3}}$^{\ddagger}$ & {\scriptsize [0.5, 3.4]} & 111.2 & +5.7 \\
claude-haiku-4-5 (2025-10-01) & 300 & 2.0 & {\scriptsize [0.9, 4.3]} & 37.0 & 0.0 & 300 & 1.0 & {\scriptsize [0.3, 2.9]} & 135.3 & +1.0 \\
meta-llama-4-maverick-17bX123e & 300 & 1.0 & {\scriptsize [0.3, 2.9]} & 92.0 & 0.0 & 300 & 0.3 & {\scriptsize [0.1, 1.9]} & 150.0 & +0.7 \\
\bottomrule
\end{tabular}
\caption{Per-model results on \emph{SCHEDBench} subset, $^{\ddagger}$ marks new results. Brackets list 95\% Wilson CIs on feasibility. Bold figures follow model feasibility through all three conditions, direct $\rightarrow$ solver-mediated $\rightarrow$ SDDL, with \textcolor{sdqwen}{blue} for qwen3.5-27b, \textcolor{sddev}{orange} for devstral-small-2-24b. $^{\dagger}$Evaluated with SDDL, not a new model.}
\label{tab:v4p-pseudomodel}
\end{table*}

\section{Results and Discussion}
We compare overall feasibility across direct natural-language scheduling (\textsc{Direct}), solver-assisted scheduling (\textsc{Solver}), and SDDL-assisted configuration (\textsc{SDDL}) for 13 models on the same 300-instance set. The main table and discussion focus on the two strongest resource-constrained \textsc{SDDL} models, while Appendix C reports additional \textsc{SDDL} results for a broader set of resource-constrained models, all of which show similar feasibility improvements. Pairwise feasibility is tested using two-sided McNemar tests on paired instance-level outcomes, with Holm correction applied jointly across  pairwise comparisons. Under \textsc{Direct}, GPT-5.5 obtains the highest feasibility at 57.0\%. Solver-assisted performance is strongly model-dependent: 10 of 13 models improve and 3 decline, with changes ranging from $-10.0\%$ to $+52.0\%$ points. Against this baseline, Qwen 3.5 27B with \textsc{SDDL} reaches 55.3\% feasibility, while Devstral Small 2 24B with \textsc{SDDL} reaches 28.3\% feasibility, both substantially improved over Direct.


\subsection{Effect on Resource-Constrained Models}
With SDDL, Qwen3.5-27B reaches 55.3\% feasibility, up significantly from 23.7\% under \textsc{Direct} generation, while Devstral-Small-2-24B similarly improves from 1.3\% to 28.3\%. For Qwen3.5-27B, generic \textsc{Solver} assistance yields only 21.7\% feasibility, slightly below \textsc{Direct}, whereas SDDL delivers gains of 31.7 and 33.7 percentage points over the two conditions, respectively ($p < 10^{-22}$).\footnote{SDDL discordant gains/losses: 100/5 and 107/6 (Qwen vs.\ DIRECT and SOLVER); 82/1 and 67/3 (Devstral).} As both solver-mediated conditions share solver and conceptual scope, these results indicate that solver access alone is insufficient, and supports SDDL’s constrained formalization-and-compilation approach. Run-failure rate also falls from 62.7\% under \textsc{Solver} to 16.0\% with \textsc{SDDL} (a 46.7-point reduction), indicating that the feasibility gain coincides with a increase in solve reliability. Devstral-Small-2-24B benefits similarly from \textsc{SDDL}, with feasibility reaching 28.3\%, vs. 7.0\% with SOLVER and 1.3\% under DIRECT ($p < 10^{-15}$), while run-failure falls from 66.7\% to 30.0\%. 


\subsection{Performance Positioning of SDDL}


We position Qwen 3.5 27B with \textsc{SDDL} against the strongest results in Table~\ref{tab:v4p-pseudomodel}. The highest direct result is 57.0\% feasibility on the 300 instance SCHEDBench subset, obtained by GPT-5.5, while the highest solver-assisted result is 56.7\%, obtained by Claude Opus 4.6. Qwen 3.5 27B with SDDL reaches 55.3\%, compared  with the generic solver-assisted results of GPT-5.5 (53.3\%) and GPT-5.4 (51.7\%), and within 1.4 and 1.7 points of the strongest solver-assisted (Claude Opus 4.6, 56.7\%) and direct (GPT-5.5, 57.0\%) configurations. SDDL closes 94.9\% of Qwen's deficit to GPT-5.5 under \textsc{Direct} (33.3 $\rightarrow$ 1.7 \%) and 96.0\% of its deficit to the strongest solver-assisted configuration (35.0 $\rightarrow$ 1.4 \%). The resulting configuration ranks among the highest feasibility overall and exceeds every generic solver-assisted configuration except Claude Opus 4.6 without increasing the capacity of the 27B model. These findings position our approach as a promising means of improving performance across resource-constrained models; with potential applicability to other resource-constrained settings.

\subsection{Optimality of Feasible Schedules}

Feasibility establishes whether a schedule satisfies the problem's constraints, but not its quality. Thus, we report the median optimality gap among feasible outputs---the difference between a schedule's objective value and the best-known value for its instance, where 0.0\% indicates that the schedule matches the best-known objective. Across all four resource-constrained models evaluated with SDDL, the median gap is 0.0\%, while feasibility ranges from 15.0\% to 55.3\%. This represents broader feasible coverage than direct generation, whose median gaps range from 111.2\% to 395.8\%, and than generic solver-code generation, which generally obtains low median gaps but solves considerably fewer instances. Qwen3.5-27B with SDDL reaches 55.3\% feasibility with a 0.0\% median gap, compared with 23.7\% feasibility and a 395.8\% median gap under \textsc{Direct}, and 21.7\% feasibility and a 2.6\% median gap under \textsc{Solver}. Since gaps are computed only over feasible outputs, the median for each condition may reflect a  different subset of instances,  and should be interpreted alongside feasibility. Still, the consistent 0.0\% median gap across SDDL configurations indicates its advantage may extend to objective quality beyond validity.


\section{Conclusion}

We introduced SDDL, a scheduling-specific language that lets models express problem structure through compact, solver-aligned primitives while delegating low-level modeling and search to a compiler and solver. On 300 SCHEDBench instances spanning multiple scheduling families, SDDL substantially raises feasibility, showing that a DSL can serve as an intermediate representation enabling resource-constrained models to match or close the deficit to the strongest evaluated configurations.

\section*{Limitations}

We do not evaluate SDDL on frontier-scale models, as our focus lies on seeking methods that improve performance for resource-constrained applications where the use of frontier scale sized models may not be possible. SDDL may also improve frontier models however is left as natural future work. Although SDDL's primitives are designed to cover a wide variety of scheduling problems, its generalization to combinatorial domains beyond scheduling remains empirically untested, and represents a natural direction for future work. Our evaluation also only covers JSSP/SM-RCPSP/MM-RCPSP scheduling families, with discrete formulations and their respective measures specified, compiler-verified, but with evaluation of their LLM-translation left to future work. Our evaluation measures single-pass, zero-shot formalization under greedy decoding. We do not evaluate iterative repair or self-correction which may recover execution failures at additional inference cost. Because models are accessed via provider APIs, temperature-0 decoding is not bitwise deterministic, and each instance is evaluated as a single draw with uncertainty quantified across instances rather than sampling seed. Our use of resource-constrained concerns only the parameter count of the language-model component and does not imply lower end-to-end compute, latency, memory, or cost for the full solver-assisted pipeline.

Based on preliminary pilot experiments, we restrict our evaluation to models with approximately 20B or more parameters. Models below this range produced substantially lower rates of valid formalizations (feas. < 5\%), making a full evaluation prohibitively uninformative under our fixed zero-shot setting. This threshold was selected empirically; therefore, our conclusions are limited to the evaluated model-size range and should not be interpreted as establishing a general minimum model size for SDDL.

\bibliography{custom}

@inproceedings{gao2023pal,
    author = {Luyu Gao and Aman Madaan and Shuyan Zhou and Uri Alon and Pengfei Liu and Yiming Yang and Jamie Callan and Graham Neubig},
    title = {{PAL}: Program-aided Language Models},
    booktitle = {Proceedings of the 40th International Conference on Machine Learning},
    series = {Proceedings of Machine Learning Research},
    volume = {202},
    pages = {10764--10799},
    year = {2023},
    publisher = {PMLR},
    url = {https://proceedings.mlr.press/v202/gao23f.html}
}

@article{2026schedbench,
  author  = {Shrenil Shaun Sharma and Avi Sharma},
  title   = {{SCHEDBench}: {A} Benchmark for Evaluating {LLM} Constraint
             Faithfulness in Natural-Language Combinatorial Scheduling},
  journal = {{arXiv}},
  volume  = {arXiv:2608.00991},
  year    = {2026},
  url     = {https://arxiv.org/abs/2608.00991},
  doi     = {10.48550/arXiv.2608.00991},
}

@article{chen2023program,
    author = {Wenhu Chen and Xueguang Ma and Xinyi Wang and William W. Cohen},
    title = {Program of Thoughts Prompting: Disentangling Computation from Reasoning for Numerical Reasoning Tasks},
    journal = {Transactions on Machine Learning Research},
    year = {2023},
    issn = {2835-8856},
    url = {https://arxiv.org/pdf/2211.12588}
}

@inproceedings{pan2023logiclm,
    author = {Liangming Pan and Alon Albalak and Xinyi Wang and William Wang},
    title = {Logic-{LM}: Empowering Large Language Models with Symbolic Solvers for Faithful Logical Reasoning},
    booktitle = {Findings of the Association for Computational Linguistics: EMNLP 2023},
    pages = {3806--3824},
    year = {2023},
    address = {Singapore},
    publisher = {Association for Computational Linguistics},
    doi = {10.18653/v1/2023.findings-emnlp.248},
    url = {https://aclanthology.org/2023.findings-emnlp.248/}
}

@inproceedings{ramamonjison2022nl4opt,
    author = {Rindranirina Ramamonjison and Timothy Yu and Raymond Li and Haley Li and Giuseppe Carenini and Bissan Ghaddar and Shiqi He and Mahdi Mostajabdaveh and Amin Banitalebi-Dehkordi and Zirui Zhou and Yong Zhang},
    title = {{NL4Opt} Competition: Formulating Optimization Problems Based on Their Natural Language Descriptions},
    booktitle = {Proceedings of the NeurIPS 2022 Competitions Track},
    series = {Proceedings of Machine Learning Research},
    volume = {220},
    pages = {189--203},
    year = {2022},
    publisher = {PMLR},
    url = {https://proceedings.mlr.press/v220/ramamonjison23a.html}
}

@inproceedings{ahmaditeshnizi2024optimus,
    author = {Ali Ahmaditeshnizi and Wenzhi Gao and Madeleine Udell},
    title = {{OptiMUS}: Scalable Optimization Modeling with ({MI}){LP} Solvers and Large Language Models},
    booktitle = {Proceedings of the 41st International Conference on Machine Learning},
    series = {Proceedings of Machine Learning Research},
    volume = {235},
    pages = {577--596},
    year = {2024},
    publisher = {PMLR},
    url = {https://proceedings.mlr.press/v235/ahmaditeshnizi24a.html}
}

@inproceedings{michailidis2024constraint,
    author = {Kostis Michailidis and Dimos Tsouros and Tias Guns},
    title = {Constraint Modelling with {LLMs} Using In-Context Learning},
    booktitle = {30th International Conference on Principles and Practice of Constraint Programming (CP 2024)},
    series = {Leibniz International Proceedings in Informatics (LIPIcs)},
    volume = {307},
    pages = {20:1--20:27},
    year = {2024},
    publisher = {Schloss Dagstuhl -- Leibniz-Zentrum f{\"u}r Informatik},
    address = {Dagstuhl, Germany},
    doi = {10.4230/LIPIcs.CP.2024.20},
    url = {https://drops.dagstuhl.de/entities/document/10.4230/LIPIcs.CP.2024.20}
}

@inproceedings{shi2025constraintllm,
    author = {Weichun Shi and Minghao Liu and Wanting Zhang and Langchen Shi and Fuqi Jia and Feifei Ma and Jian Zhang},
    title = {{ConstraintLLM}: A Neuro-Symbolic Framework for Industrial-Level Constraint Programming},
    booktitle = {Proceedings of the 2025 Conference on Empirical Methods in Natural Language Processing},
    pages = {15999--16019},
    year = {2025},
    address = {Suzhou, China},
    publisher = {Association for Computational Linguistics},
    doi = {10.18653/v1/2025.emnlp-main.809},
    url = {https://aclanthology.org/2025.emnlp-main.809/}
}

@inproceedings{jiang-etal-2026-reasoning-combinatorial,
    title = {Reasoning in a Combinatorial and Constrained World: Benchmarking {LLM}s on Natural-Language Combinatorial Optimization},
    author = {Jiang, Xia and
      Chen, Jing and
      Zhang, Cong and
      Gao, Jie and
      Hu, Chengpeng and
      Zhang, Chenhao and
      Wu, Yaoxin and
      Zhang, Yingqian},
    editor = {Liakata, Maria and
      Moreira, Viviane P. and
      Zhang, Jiajun and
      Jurgens, David},
    booktitle = {Findings of the Association for Computational Linguistics: ACL 2026},
    month = jul,
    year = {2026},
    address = {San Diego, California, United States},
    publisher = {Association for Computational Linguistics},
    url = {https://aclanthology.org/2026.findings-acl.1529/},
    doi = {10.18653/v1/2026.findings-acl.1529},
    pages = {30592--30648},
    isbn = {979-8-89176-395-1}
}

@inproceedings{yin2018tranx,
    author = {Pengcheng Yin and Graham Neubig},
    title = {{TRANX}: A Transition-based Neural Abstract Syntax Parser for Semantic Parsing and Code Generation},
    booktitle = {Proceedings of the 2018 Conference on Empirical Methods in Natural Language Processing: System Demonstrations},
    pages = {7--12},
    year = {2018},
    address = {Brussels, Belgium},
    publisher = {Association for Computational Linguistics},
    doi = {10.18653/v1/D18-2002},
    url = {https://aclanthology.org/D18-2002/}
}

@inproceedings{geng2023grammar,
    author = {Saibo Geng and Martin Josifoski and Maxime Peyrard and Robert West},
    title = {Grammar-Constrained Decoding for Structured {NLP} Tasks without Finetuning},
    booktitle = {Proceedings of the 2023 Conference on Empirical Methods in Natural Language Processing},
    pages = {10932--10952},
    year = {2023},
    address = {Singapore},
    publisher = {Association for Computational Linguistics},
    doi = {10.18653/v1/2023.emnlp-main.674},
    url = {https://aclanthology.org/2023.emnlp-main.674/}
}

@misc{abgaryan2024llmsschedule,
    author = {Henrik Abgaryan and Ararat Harutyunyan and Tristan Cazenave},
    title = {{LLMs} can Schedule},
    year = {2024},
    eprint = {2408.06993},
    archivePrefix = {arXiv},
    primaryClass = {cs.AI},
    url = {https://arxiv.org/abs/2408.06993}
}

@misc{abgaryan2025starjob,
    author = {Henrik Abgaryan and Tristan Cazenave and Ararat Harutyunyan},
    title = {Starjob: Dataset for {LLM}-Driven Job Shop Scheduling},
    year = {2025},
    eprint = {2503.01877},
    archivePrefix = {arXiv},
    journal = {arXiv},
    primaryClass = {cs.LG},
    url = {https://arxiv.org/abs/2503.01877}
}

@inproceedings{liao-etal-2026-nl,
    title = {{NL} $\Rightarrow$ Schedule: Evaluate Multitask Scheduling Capability of Large Language Models},
    author = {Liao, Wenrui and
      Du, Weihong and
      Li, Yi and
      Liang, Hongru and
      Lei, Wenqiang},
    editor = {Liakata, Maria and
      Moreira, Viviane P. and
      Zhang, Jiajun and
      Jurgens, David},
    booktitle = {Proceedings of the 64th Annual Meeting of the Association for Computational Linguistics (Volume 1: Long Papers)},
    month = jul,
    year = {2026},
    address = {San Diego, California, United States},
    publisher = {Association for Computational Linguistics},
    url = {https://aclanthology.org/2026.acl-long.1648/},
    doi = {10.18653/v1/2026.acl-long.1648},
    pages = {35620--35640},
    isbn = {979-8-89176-390-6}
}

@misc{jain2025rconstraintbench,
    author = {Raj Jain and Marc Wetter},
    title = {{R-ConstraintBench}: Evaluating {LLM}s on {NP}-Complete Scheduling},
    year = {2025},
    eprint = {2508.15204},
    archivePrefix = {arXiv},
    primaryClass = {cs.AI},
    url = {https://arxiv.org/abs/2508.15204}
}

@inproceedings{tam2024let,
    author = {Zhi Rui Tam and Cheng-Kuang Wu and Yi-Lin Tsai and Chieh-Yen Lin and Hung-yi Lee and Yun-Nung Chen},
    title = {Let Me Speak Freely? {A} Study on the Impact of Format Restrictions on Large Language Model Performance},
    booktitle = {Proceedings of the 2024 Conference on Empirical Methods in Natural Language Processing: Industry Track},
    pages = {1218--1236},
    year = {2024},
    address = {Miami, Florida, US},
    publisher = {Association for Computational Linguistics},
    doi = {10.18653/v1/2024.emnlp-industry.91},
    url = {https://aclanthology.org/2024.emnlp-industry.91/}
}

@inproceedings{valmeekam2023planbench,
    author = {Karthik Valmeekam and Matthew Marquez and Alberto Olmo and Sarath Sreedharan and Subbarao Kambhampati},
    title = {{PlanBench}: An Extensible Benchmark for Evaluating Large Language Models on Planning and Reasoning about Change},
    booktitle = {Advances in Neural Information Processing Systems},
    volume = {36},
    pages = {38975--38987},
    year = {2023},
    publisher = {Curran Associates, Inc.},
    doi = {10.52202/075280-1693},
    url = {https://proceedings.neurips.cc/paper_files/paper/2023/hash/7a92bcdede88c7afd108072faf5485c8-Abstract-Datasets_and_Benchmarks.html}
}

@inproceedings{kambhampati2024llms,
    author = {Subbarao Kambhampati and Karthik Valmeekam and Lin Guan and Mudit Verma and Kaya Stechly and Siddhant Bhambri and Lucas Paul Saldyt and Anil B. Murthy},
    title = {Position: {LLM}s Can't Plan, But Can Help Planning in {LLM}-Modulo Frameworks},
    booktitle = {Proceedings of the 41st International Conference on Machine Learning},
    series = {Proceedings of Machine Learning Research},
    volume = {235},
    pages = {22895--22907},
    year = {2024},
    publisher = {PMLR},
    url = {https://proceedings.mlr.press/v235/kambhampati24a.html}
}

@inproceedings{yao2024collie,
    author = {Shunyu Yao and Howard Chen and Austin W. Hanjie and Runzhe Yang and Karthik Narasimhan},
    title = {{COLLIE}: Systematic Construction of Constrained Text Generation Tasks},
    booktitle = {The Twelfth International Conference on Learning Representations},
    year = {2024},
    url = {https://openreview.net/forum?id=kxgSlyirUZ}
}

@inproceedings{kesseli2025logicpy,
    author = {Pascal Kesseli and Peter O'Hearn and Ricardo S. Cabral},
    title = {{Logic.py}: Bridging the Gap between {LLM}s and Constraint Solvers},
    booktitle = {Advances in Neural Information Processing Systems},
    volume = {38},
    year = {2025},
    publisher = {Curran Associates, Inc.},
    url = {https://proceedings.neurips.cc/paper_files/paper/2025/hash/5a29c3d172b80bab1238ddc227246c52-Abstract-Conference.html}
}

\appendix

\section{SDDL Semantics and Compilation}

\subsection{Primitive and Property Reference}
\label{sec:app-primitives}
Table~\ref{tab:app-primitives} lists every primitive, accepted property, and
well-formedness requirement; Table~\ref{tab:app-measures} defines the five
registered penalty measures.

\begin{table}[h]
\centering\scriptsize
\setlength{\tabcolsep}{3pt}
\begin{tabular}{@{}p{2.0cm}p{1.5cm}p{3.6cm}@{}}
\toprule
\textbf{Measure} & \textbf{Archetype} & \textbf{Definition (minimized)} \\
\midrule
\texttt{makespan} & continuous & $\max_i \mathrm{end}_i$ over all scheduled tasks \\
\texttt{capacity} & discrete & total enrolment exceeding room capacity, summed over assignments \\
\texttt{spread} & discrete & shortfall below each task's \texttt{min\_days} distinct meeting days \\
\texttt{isolated} & discrete & count of meetings with no adjacent same-group meeting \\
\texttt{room\_stability} & discrete & number of distinct rooms used by a task beyond the first \\
\bottomrule
\end{tabular}
\caption{The five registered penalty measures. Each is implemented once in
the compiler and reused across programs; \texttt{makespan} is the only
measure exercised by the evaluated families.}
\label{tab:app-measures}
\end{table}

\subsection{Compiler-Lowering Summary}
\label{sec:app-lowering}
Table~\ref{tab:app-lowering} maps each SDDL construct to the CP-SAT encoding
the compiler emits; all solver variables are introduced by the compiler.

\subsection{Correctness Arguments}
\label{app:correctness}
Adequacy is proven statement-by-statement; the constructs whose
lowerings introduce auxiliary variables require the following arguments:
\begin{itemize}\itemsep2pt
\item \textbf{Cumulative resources.} \texttt{AddCumulative} enforces
$\sum_{i:\, s_i \leq t < e_i} d_{ir} \leq k$ at every $t$, the reference
condition. Multi-mode tasks contribute one \emph{optional} interval per mode,
present iff its mode Boolean holds; \texttt{AddExactlyOne} presents exactly
the selected mode's interval and demand, so solutions correspond one-to-one
with reference schedules.
\item \textbf{Mode selection.} \texttt{AddExactlyOne}$(b_{i\cdot})$ makes
selection total and unique; channelling $b_{im} \Rightarrow d_i = d_{im}$
fixes the master interval's duration, per-mode optional intervals share the
task's start, and nonrenewable consumption $\sum_m c_{imr} b_{im}$ equals the
selected mode's consumption. Starts, durations, consumptions are preserved in both directions.
\item \textbf{Reified penalty measures.} Every measure lowers to variables
constrained to \emph{equal} the measured quantity
(\texttt{AddMaxEquality} for makespan; complementary
\texttt{OnlyEnforceIf} pairs, $b \Leftrightarrow$ condition, for discrete
counts), never one-sided bounds a minimizer could exploit; compiled
objectives therefore equal reference objectives on all feasible schedules.
\item \textbf{Horizon soundness.} With
$H = T + \sum_i \max_{m \in M_i} d_{im}$ ($T$ = one plus the latest
forbidden time; zero for the evaluated families), the serial schedule in
topological order is feasible with span $\leq H$, so truncation to $[0,H]$
never empties the feasible set. The unrestricted optimum is at most the
serial span $\leq H$, and any schedule attaining it has every end within
$[0,H]$; the compiled optimum equals the reference optimum.
\item \textbf{Composition.} Lowerings share only the task variables
$(s_i, d_i, e_i, b_{im})$ whose meaning the items above fix, so per-statement
adequacy composes to program adequacy; the residual failure surface is the
model-produced formalization, measured by the independent verifier
(\ref{sec:app-verification}).
\end{itemize}

\subsection{Worked Translation Example}
\label{sec:app-worked}
Figure~\ref{fig:app-worked} shows one complete translation. CP-SAT returns
the optimum, makespan 7 (welding: \texttt{j0\_o0} $[0,3)$,
\texttt{j1\_o1} $[4,7)$; grinding: \texttt{j1\_o0} $[0,4)$,
\texttt{j0\_o1} $[4,6)$; the lower bound from Batch Concord's $4{+}3$
chain), rendered back to schedule lines via \texttt{label}/\texttt{position}.

\subsection{Full SDDL, Transpiler, Compiler}
\label{sec:app-artifact}
\begin{itemize}\itemsep1pt
\item literal-only \texttt{ast}-based parser ($\sim$170 lines; generated
text is never executed);
\item CP-SAT transpiler ($\sim$560 lines; both archetypes, all five
measures);
\item renderer mapping solved variables to schedule lines via task labels;
\item evaluation harness, per-instance outputs, and scoring records.
\end{itemize}

\section{Experimental Details}

\subsection{SCHEDBench Evaluation Subset}
\label{sec:app-subset}
Instances chosen randomly using a fixed subsampling seed 42; instance identifiers are the
\texttt{source\_instance} fields of the benchmark. Gaps are
computed against canonical optimum or best-known solution; and breakdown of per family subset composition is below.

\begin{table}[h]
\centering\scriptsize
\setlength{\tabcolsep}{3.5pt}
\begin{tabular}{@{}llr@{}}
\toprule
\textbf{Family} & \textbf{Source suites} & $N$ \\
\midrule
JSSP & Taillard, DMU, LA, ORB, ABZ, SWV & 100 \\
SM-RCPSP & PSPLIB J30--J120 & 100 \\
MM-RCPSP & PSPLIB MM J10--J30 & 100 \\
\bottomrule
\end{tabular}
\caption{Evaluation-subset composition.}
\label{tab:app-subset}
\end{table}

\subsection{Prompt Templates}
\label{sec:app-prompts}
Figures~\ref{fig:app-prompt-direct} and~\ref{fig:app-prompt-solver} reproduce the
three fixed, instance-independent system prompts; the user message is the
instance's problem text plus its response-format section, identical across
tested conditions. All conditions are evaluated zero-shot, without worked
examples.

\subsubsection{Direct Generation}
Figure~\ref{fig:app-prompt-direct} reproduces the \textsc{Direct} system
prompt.

\subsubsection{Generic Solver-Code Generation}
Figure~\ref{fig:app-prompt-solver} reproduces the \textsc{Solver} system
prompt.

\subsubsection{SDDL Generation}
\label{sec:app-prompt-sddl}
The \textsc{SDDL} system prompt is reproduced beginning on
p.~\pageref{lst:app-prompt-sddl}.

\subsection{Model, Inference, and Solver Configuration}
\label{model}
\begin{table}[h]
\centering\scriptsize
\setlength{\tabcolsep}{3pt}
\begin{tabular}{@{}p{2.2cm}p{5.0cm}@{}}
\toprule
\textbf{Setting} & \textbf{Value} \\
\midrule
Resource-Constrained models & qwen3.5-27b (2026-02-24), devstral-small-2-24b (2025-12-09), qwen3-coder-30b-a3b (2025-07-31), magistral-small-24b (2025-09-17) \\
Reproduced & pinned versions of Table~1 (SCHEDBench) \\
Decoding & greedy (temperature 0), zero-shot, single pass, no tools \\
Max output tokens & 96{,}000 default; provider ceilings; token-terminated responses retained and scored \\
Sandbox & no network, fresh directory, 4\,GB memory, 300\,s wall-clock limit; CP-SAT budget fixed at 240\,s in both solver-mediated conditions \\
Software & Python 3.11.2; OR-Tools 9.15.6755 (CP-SAT, default parameters except \texttt{max\_time\_in\_seconds}) \\
\bottomrule
\end{tabular}
\caption{Inference, sandbox, and solver configuration.}
\label{tab:app-config}
\end{table}

\subsection{Verification and Outcome Accounting}
\label{sec:app-verification}
The verifier parses emitted schedule lines and re-derives feasibility and
the objective directly from the canonical source instance; it shares no
code with the DSL parser, transpiler, or CP-SAT, and is the same component
that scores \textsc{Direct} (whose pipeline involves no compiler). Solver
status is never trusted.
\begin{itemize}\itemsep2pt
\item \textbf{Run-fail}: no scoreable schedule (unparseable/empty output,
parse rejection, transpile or runtime error, sandbox timeout, or no
completed generation).
\item \textbf{No-solution}: program executed; solver reported infeasibility
or returned nothing within budget.
\item \textbf{Infeasible}: schedule produced but violates a hard constraint
(classified precedence / machine-overlap / resource-capacity, plus coverage
for missing activities).
\item \textbf{Feasible}: all hard constraints verified; objective
recomputed from the schedule.
\item \textbf{\textsc{Direct} accounting}: unparseable $\to$ run-fail;
parsed-but-violating $\to$ infeasible; no-solution cannot occur.
\item \textbf{Denominator}: $N{=}300$ per configuration; instances without
a completed generation count as run-fail, so each row's outcomes sum to
100\%.
\item \textbf{Wilson 95\% CI}:
$\big(\hat p + \tfrac{z^2}{2n} \pm z\sqrt{\tfrac{\hat p(1-\hat p)}{n} +
\tfrac{z^2}{4n^2}}\big)\big/\big(1+\tfrac{z^2}{n}\big)$, $z=1.96$.
\item \textbf{Paired tests}: two-sided exact McNemar on instance-level
feasibility, $p = \min\!\big(1, 2\sum_{i \leq \min(b,c)} \binom{n}{i}
2^{-n}\big)$, $n = b+c$; Holm correction applied jointly across all
reported pairwise comparisons; unscored instances count as not-feasible on
both sides (no verdict changes on the jointly-scored subset).
Table~\ref{tab:app-mcnemar} lists all discordant counts.
\end{itemize}

\section{Supplemental Results}



\subsection{Error Analysis}
\label{sec:app-error-analysis}
Representative exemplars (one per dominant code):
\begin{itemize}\itemsep2pt
\item \textbf{Dropped edge}: all 32 tasks, 4 resources, and every
duration/demand of a 30-activity MM-RCPSP instance transcribed correctly;
two \texttt{before()} entries omitted.
\item \textbf{Conflation}: instance contains both ``Waste Removal
Planning'' and ``North Waste Removal Planning''; the program declares only
\texttt{north\_waste\_removal\_planning} yet writes
\texttt{before("waste\_removal\_planning", ...)} $\to$ \texttt{KeyError}.
\item \textbf{Derailment}: correct DSL for 160 lines, then drift into
natural-language commentary; rejected at parse time.
\item \textbf{Duplicate declarations}: a job-shop program
emits its 300 operation declarations twice
(600 task() calls), leaving edge semantics
attached to shadowed duplicates.
\end{itemize}

\clearpage
\begin{table*}[p]
\centering\footnotesize
\setlength{\tabcolsep}{4pt}
\begin{tabular}{@{}p{2.5cm}p{1.5cm}p{1.7cm}p{8.9cm}@{}}
\toprule
\textbf{Primitive} & \textbf{Property} & \textbf{Value} & \textbf{Meaning / well-formedness} \\
\midrule
\texttt{task(id, ...)} & \texttt{duration} & int $\geq 0$ & processing time; declares a \emph{continuous} task \\
 & \texttt{modes} & list of maps & alternative execution modes, each a map with \texttt{duration} and optional \texttt{demands}/\texttt{consumes}; declares a continuous multi-mode task; mutually exclusive with a top-level \texttt{duration} \\
 & \texttt{demands} & map $\{$rid: int$\}$ & renewable units held while active; keys must name declared \texttt{resource()} ids with \texttt{capacity} \\
 & \texttt{consumes} & map $\{$rid: int$\}$ & nonrenewable units expended once; keys must name resources with \texttt{total} \\
 & \texttt{machine} & string rid & fixed disjunctive-resource assignment; pair with \texttt{no\_overlap(rid)} \\
 & \texttt{job}, \texttt{position} & int, int & job index and 0-based operation index (job-shop bookkeeping used by the renderer) \\
 & \texttt{label} & string & verbatim display name from the problem text; consumed by the renderer/verifier, not the solver \\
 & \texttt{count} & int $\geq 1$ & number of meetings; declares a \emph{discrete} (timetabling) task \\
 & \texttt{min\_days}, \texttt{students} & int & discrete-archetype spread/enrolment attributes \\
 & \texttt{demand} & int $\geq 1$ & units required while assigned (discrete archetype: enrolment, checked against room \texttt{capacity} and scored by the \texttt{capacity} measure) \\
\texttt{resource(id, ...)} & \texttt{capacity} & int $\geq 1$ & renewable per-time capacity $\Rightarrow$ cumulative semantics; for the discrete archetype, a room's seat capacity \\
 & \texttt{total} & int $\geq 1$ & nonrenewable project-wide budget $\Rightarrow$ linear budget semantics \\
 & (neither) & --- & strictly disjunctive resource; meaningful with \texttt{no\_overlap} \\
\texttt{before(a, b)} & --- & task ids & hard finish-to-start precedence: $\mathrm{end}_a \leq \mathrm{start}_b$; both ids must be declared \\
\texttt{no\_overlap(r)} & --- & resource id & at most one assigned task active on $r$ at any time \\
\texttt{conflict(...)} & \texttt{group} & string (opt.) & listed tasks may not occupy the same time slot; group label also keys group-level objective terms (discrete archetype) \\
\texttt{not\_at(id, ...)} & \texttt{day}, \texttt{period} & int & forbids execution at the given absolute time; on continuous tasks the forbidden times enter the horizon offset $T$ (\S3.7) \\
\texttt{penalize(m, ...)} & \texttt{weight} & int $\geq 1$ & soft objective term; $m$ must be a registered measure name (Table~\ref{tab:app-measures}) \\
\bottomrule
\end{tabular}
\caption{Complete primitive and property reference. A well-formed
\texttt{task} satisfies exactly one archetype rule: it is \emph{continuous}
iff it specifies \texttt{duration} or \texttt{modes} and no \texttt{count},
and \emph{discrete} iff it specifies \texttt{count}; archetype assignment is
therefore deterministic and complete, and a task specifying both (or neither) is rejected as malformed.}
\label{tab:app-primitives}
\end{table*}

\begin{table*}[p]
\centering\footnotesize
\setlength{\tabcolsep}{5pt}
\begin{tabular}{@{}p{4.4cm}p{10.7cm}@{}}
\toprule
\textbf{SDDL construct} & \textbf{CP-SAT lowering} \\
\midrule
continuous \texttt{task} (fixed duration $d$) & $s_i, e_i \in [0, H]$ (\texttt{NewIntVar}); interval \texttt{NewIntervalVar}$(s_i, d, e_i)$ \\
continuous \texttt{task} with \texttt{modes} $M_i$ & mode Booleans $b_{im}$ (\texttt{NewBoolVar}) with \texttt{AddExactlyOne}; duration variable $d_i \in [\min_m d_{im}, \max_m d_{im}]$ channelled by $b_{im} \Rightarrow d_i = d_{im}$ (\texttt{OnlyEnforceIf}); a master interval on $(s_i, d_i, e_i)$ plus one \texttt{NewOptionalFixedSizeIntervalVar}$(s_i, d_{im}, b_{im})$ per mode \\
\texttt{before(a, b)} & \texttt{model.Add}$(s_b \geq e_a)$ \\
\texttt{no\_overlap(r)} & \texttt{AddNoOverlap} over the intervals of tasks with \texttt{machine} $= r$ \\
renewable resource ($\texttt{capacity}=k$) & \texttt{AddCumulative}(intervals, demands, $k$); multi-mode tasks contribute their per-mode \emph{optional} intervals with mode-specific demands \\
nonrenewable resource ($\texttt{total}=B$) & linear budget $\sum_i \sum_m c_{im}\, b_{im} \leq B$ (fixed-mode consumption enters as constants) \\
\texttt{conflict(...)} & \texttt{AddAllDifferent} over the members' slot variables (discrete archetype) \\
discrete \texttt{task} (\texttt{count} $c$) & $c$ meetings, each with day, period, and room integer variables and channelled slot $= \mathrm{day} \cdot N_P + \mathrm{period}$; room assignment respects seat capacity \\
\texttt{not\_at(id, day, period)} & continuous tasks: reified exclusion $e_i \leq t \,\lor\, s_i > t$ per forbidden time $t$; discrete tasks: slot-exclusion constraints on the day/period variables, reified per meeting \\
\texttt{penalize("makespan", w)} & $\mathit{ms} \in [0,H]$ with \texttt{AddMaxEquality}$(\mathit{ms}, \{e_i\})$; contributes $w \cdot \mathit{ms}$ \\
objective & \texttt{Minimize}$\big(\sum \text{penalty terms}\big)$ \\
\bottomrule
\end{tabular}
\caption{Statement-by-statement lowering into CP-SAT. Discrete-archetype
tasks lower to day, period, and room integer variables with a channelled slot
variable $\mathrm{slot} = \mathrm{day} \cdot N_P + \mathrm{period}$.}
\label{tab:app-lowering}
\end{table*}

\begin{figure*}[p]
\textbf{Natural-language statement.} \emph{2 jobs, 2 machines. Job ``Batch
Northview'': step 1 on Main Welding Station for 3 min, step 2 on West
Grinding Station for 2 min. Job ``Batch Concord'': step 1 on West Grinding
Station for 4 min, step 2 on Main Welding Station for 3 min.}

\textbf{SDDL program (model output):}

\begin{lstlisting}[basicstyle=\ttfamily\scriptsize,breaklines=true,breakindent=0pt,columns=fullflexible,frame=single,framesep=3pt,xleftmargin=2pt,xrightmargin=2pt]
resource("main_welding_station")
resource("west_grinding_station")
task("j0_o0", label="Batch Northview", job=0, position=0, machine="main_welding_station", duration=3)
task("j0_o1", label="Batch Northview", job=0, position=1, machine="west_grinding_station", duration=2)
task("j1_o0", label="Batch Concord", job=1, position=0, machine="west_grinding_station", duration=4)
task("j1_o1", label="Batch Concord", job=1, position=1, machine="main_welding_station", duration=3)
before("j0_o0", "j0_o1")
before("j1_o0", "j1_o1")
no_overlap("main_welding_station")
no_overlap("west_grinding_station")
penalize("makespan", weight=1)
\end{lstlisting}

\textbf{Principal compiled constraints:}

\begin{lstlisting}[basicstyle=\ttfamily\scriptsize,breaklines=true,breakindent=0pt,columns=fullflexible,frame=single,framesep=3pt,xleftmargin=2pt,xrightmargin=2pt]
horizon = 12
starts['j0_o0'] = model.NewIntVar(0, horizon, 'j0_o0_s')
ends['j0_o0']   = model.NewIntVar(0, horizon, 'j0_o0_e')
intervals['j0_o0'] = model.NewIntervalVar(starts['j0_o0'], 3, ends['j0_o0'], 'j0_o0_iv')
...                                  # same three lines per task
model.Add(starts['j0_o1'] >= ends['j0_o0'])
model.Add(starts['j1_o1'] >= ends['j1_o0'])
model.AddNoOverlap([intervals['j0_o0'], intervals['j1_o1']])  # main_welding_station
model.AddNoOverlap([intervals['j0_o1'], intervals['j1_o0']])  # west_grinding_station
_ms = model.NewIntVar(0, horizon, 'makespan')
model.AddMaxEquality(_ms, [ends['j0_o0'], ends['j0_o1'], ends['j1_o0'], ends['j1_o1']])
penalties.append(_ms)
total = model.NewIntVar(0, 99999999, 'total')
model.Add(total == sum(penalties))
model.Minimize(total)
\end{lstlisting}

\caption{Worked end-to-end translation: natural-language instance
$\rightarrow$ SDDL $\rightarrow$ compiled CP-SAT model.}
\label{fig:app-worked}
\end{figure*}

\begin{figure*}[p]
\begin{lstlisting}[basicstyle=\ttfamily\scriptsize,breaklines=true,breakindent=0pt,columns=fullflexible,frame=single,framesep=3pt,xleftmargin=2pt,xrightmargin=2pt]
You are completing an automated benchmark. The user message contains a scheduling problem and an exact output format. Begin your response with the first schedule line and output only the schedule in that format -- nothing else: no preamble, no explanation, no reasoning, no markdown formatting (no **bold**, no `inline code`, no triple-backtick code blocks), no XML tags such as <schedule> or <answer>, no tool tags, no Python code. Do not call any tools, do not invoke external solvers, and do not generate code to be executed -- solve the problem yourself using only your own reasoning. Your response is fed directly into a parser; any extra characters cause the response to be discarded.
\end{lstlisting}
\caption{\textsc{Direct} system prompt.}
\label{fig:app-prompt-direct}
\end{figure*}

\begin{figure*}[p]
\begin{lstlisting}[basicstyle=\ttfamily\scriptsize,breaklines=true,breakindent=0pt,columns=fullflexible,frame=single,framesep=3pt,xleftmargin=2pt,xrightmargin=2pt]
You are completing an automated benchmark. The user message contains a scheduling problem written in natural language and an exact output format. Do not solve the problem in your head and do not emit a schedule directly. Instead, write a single self-contained Python 3 program that solves the problem using ortools.sat.python.cp_model and prints the resulting schedule to standard output in exactly the output format the user message specifies. The first character of your response must be the first character of the program (typically "from" or "import"); output nothing else -- no preamble, no explanation, no reasoning, no markdown formatting, no triple-backtick code blocks, no XML tags, no comments outside the program body.

Every piece of instance data must be transcribed from the user message into the program as Python literals. Do not read from any file, do not import a data module, do not make any network request, and do not depend on any environment variable, command-line argument, or side channel. The program must run to completion on its own with `python program.py` in a fresh directory with no network access. Allowed imports are the Python standard library and ortools.sat.python.cp_model -- nothing else.

The program's standard output must be, and only be, schedule lines in exactly the format the user message specifies -- the same character-level format that the direct-schedule condition uses. Do not print status labels, diagnostics, timings, or objective values; do not wrap the output in any envelope or code fence. If the solver returns INFEASIBLE or UNKNOWN, print nothing and exit cleanly. The program's standard output is fed unchanged into the same parser used to score direct-schedule outputs; any extra characters cause the response to be discarded.

## CP-SAT modeling guide

model = cp_model.CpModel(). Give every operation (job shop) or activity
(project scheduling) one interval: start and end as NewIntVar bounded by
horizon = sum of all durations, linked by
NewIntervalVar(start, duration, end). Milestones are intervals of
duration 0.

## Constraints

- Precedence: for each successor entry, model.Add(start_b >= end_a).
- Job-shop machine (one operation at a time): AddNoOverlap over exactly the
  intervals assigned to that machine.
- Renewable resource pool (capacity limit while active): AddCumulative over
  the demanding intervals with their per-task demands and the pool's
  capacity. NEVER encode a renewable pool with AddNoOverlap.
- Nonrenewable budget (consumed once): a single linear constraint,
  model.Add(sum of consumed amounts <= total).
- Alternative execution modes: one BoolVar per mode with AddExactlyOne;
  each mode contributes NewOptionalIntervalVar carrying that mode's
  duration, its demands to the relevant AddCumulative calls, and its
  consumption to the budget constraint.
- Slot-conflict groups (timetabling): AddAllDifferent over the tasks' slot
  variables; forbidden times: model.Add(slot != t).

## Objective

makespan = NewIntVar(0, horizon); AddMaxEquality(makespan, all ends);
model.Minimize(makespan). Solve with cp_model.CpSolver(); set
solver.parameters.max_time_in_seconds = 240.

## Rules

1. One interval per operation or activity, including start and finish
   milestones.
2. One AddNoOverlap per job-shop machine; AddCumulative for every
   capacity-limited pool.
3. One precedence constraint per successor entry; transcribe all of an
   activity's listed successors before moving to the next. Job-shop steps
   are chained consecutively.
4. An activity with one fixed profile uses a plain interval, not modes.
5. Transcribe every duration, demand, capacity, and total exactly as given.
6. Minimize makespan.
7. Print ONLY schedule lines, with display names exactly as written in the
   problem text.\end{lstlisting}
\caption{\textsc{Solver} system prompt.}
\label{fig:app-prompt-solver}
\end{figure*}

\begin{table*}[p]
\centering\footnotesize
\begin{tabular}{@{}llrrl@{}}
\toprule
\textbf{Model} & \textbf{Contrast} & $b$ & $c$ & $p$ \\
\midrule
qwen3.5-27b & \textsc{SDDL} vs.\ \textsc{Direct} & 100 & 5 & $5.0\times10^{-24}$ \\
            & \textsc{SDDL} vs.\ \textsc{Solver} & 107 & 6 & $5.2\times10^{-25}$ \\
devstral-small-2-24b & \textsc{SDDL} vs.\ \textsc{Direct} & 82 & 1 & $1.7\times10^{-23}$ \\
            & \textsc{SDDL} vs.\ \textsc{Solver} & 67 & 3 & $9.7\times10^{-17}$ \\
qwen3-coder-30b-a3b & \textsc{SDDL} vs.\ \textsc{Direct} & 70 & 1 & $6.1\times10^{-20}$ \\
            & \textsc{SDDL} vs.\ \textsc{Solver} & 61 & 2 & $4.4\times10^{-16}$ \\
magistral-small-24b & \textsc{SDDL} vs.\ \textsc{Direct} & 41 & 0 & $9.1\times10^{-13}$ \\
            & \textsc{SDDL} vs.\ \textsc{Solver} & 42 & 5 & $2.5\times10^{-8}$ \\
\bottomrule
\end{tabular}
\caption{McNemar discordant counts ($b$: \textsc{SDDL}-only feasible;
$c$: comparison-only feasible) and two-sided $p$-values for the within-model
contrasts; all remain significant after Holm correction ($p_{\mathrm{adj}} < 10^{-6}$).}
\label{tab:app-mcnemar}
\end{table*}


\definecolor{sdqwen}{HTML}{0F62FE}
\definecolor{sddev}{HTML}{D9480F}
\definecolor{sdqc}{HTML}{2B8A3E}
\definecolor{sdmag}{HTML}{9C36B5}
\begin{table*}[t]
\centering
\footnotesize
\setlength{\tabcolsep}{2pt}
\begin{tabular}{@{}l rrcr@{\hspace{3pt}}r @{\hspace{14pt}} rrcr @{\hspace{10pt}} r@{}}
\toprule
& \multicolumn{5}{c@{\hspace{14pt}}}{\textbf{Solver-Mediated Generation}} & \multicolumn{4}{c@{\hspace{10pt}}}{\textbf{Direct Generation}} & \\
\cmidrule(lr){2-6} \cmidrule(lr){7-10}
\textbf{Model} & $N$ & {\shortstack{\textbf{Feas.}\\(\%)}} & {\shortstack{\textbf{95\% CI}\\(\%)}} & {\shortstack{\textbf{R.\,fail}\\(\%)}} & {\shortstack{\textbf{Med.\ gap}\\(\%)}} & $N$ & {\shortstack{\textbf{Feas.}\\(\%)}} & {\shortstack{\textbf{95\% CI}\\(\%)}} & {\shortstack{\textbf{Med.\ gap}\\(\%)}} & {\shortstack{\textbf{$\Delta$\ Feas.}\\{}}} \\
\midrule
\textbf{qwen3.5-27b $+$ SDDL} & 300 & \textcolor{sdqwen}{\textbf{55.3}} & {\scriptsize [49.7, 60.9]} & 16.0 & 0.0 & 300 & \textcolor{sdqwen}{\textbf{23.7}} & {\scriptsize [19.2, 28.8]} & 395.8 & \textbf{+31.7} \\
\textbf{devstral-small-2-24b $+$ SDDL} & 300 & \textcolor{sddev}{\textbf{28.3}} & {\scriptsize [23.5, 33.7]} & 30.0 & 0.0 & 300 & \textcolor{sddev}{\textbf{1.3}} & {\scriptsize [0.5, 3.4]} & 111.2 & +27.0 \\
\textbf{qwen3-coder-30b-a3b $+$ SDDL} & 300 & \textcolor{sdqc}{\textbf{23.3}} & {\scriptsize [18.9, 28.4]} & 28.7 & 0.0 & 300 & \textcolor{sdqc}{\textbf{0.3}} & {\scriptsize [0.1, 1.9]} & 357.1 & +23.0 \\
qwen/qwen3.5-27b (2026-02-24) & 300 & \textcolor{sdqwen}{\textbf{21.7}} & {\scriptsize [17.4, 26.7]} & 62.7 & 2.6 & 300 & \textcolor{sdqwen}{\textbf{23.7}} & {\scriptsize [19.2, 28.8]} & 395.8 & $-$2.0 \\
\textbf{magistral-small-24b $+$ SDDL} & 300 & \textcolor{sdmag}{\textbf{15.0}} & {\scriptsize [11.4, 19.5]} & 47.3 & 0.0 & 300 & \textcolor{sdmag}{\textbf{1.3}} & {\scriptsize [0.5, 3.4]} & 159.9 & +13.7 \\
devstral-small-2-24b (2025-12-09) & 300 & \textcolor{sddev}{\textbf{7.0}} & {\scriptsize [4.6, 10.5]} & 66.7 & 0.0 & 300 & \textcolor{sddev}{\textbf{1.3}} & {\scriptsize [0.5, 3.4]} & 111.2 & +5.7 \\
qwen3-coder-30b-a3b (2025-07-31) & 300 & \textcolor{sdqc}{\textbf{3.7}} & {\scriptsize [2.1, 6.4]} & 74.3 & 0.0 & 300 & \textcolor{sdqc}{\textbf{0.3}} & {\scriptsize [0.1, 1.9]} & 357.1 & +3.3 \\
magistral-small-24b (2025-09-17) & 300 & \textcolor{sdmag}{\textbf{2.7}} & {\scriptsize [1.4, 5.2]} & 88.0 & 0.0 & 300 & \textcolor{sdmag}{\textbf{1.3}} & {\scriptsize [0.5, 3.4]} & 159.9 & +1.3 \\
\bottomrule
\end{tabular}
\caption{Resource-constrained models across all three conditions, direct $\rightarrow$ solver-mediated $\rightarrow$ SDDL, in the format of Table~\ref{tab:v4p-pseudomodel}. Brackets list 95\% Wilson CIs on feasibility. Color follows one model across rows, with \textcolor{sdqwen}{blue} for qwen3.5-27b, \textcolor{sddev}{orange} for devstral-small-2-24b, \textcolor{sdqc}{green} for qwen3-coder-30b-a3b, \textcolor{sdmag}{purple} for magistral-small-24b.}
\label{tab:v4p-small-models}
\end{table*}

\begin{table*}[p]
\centering\footnotesize
\begin{tabular}{@{}llrrrr@{}}
\toprule
\textbf{Cond.} & \textbf{Family} & $N$ & \textbf{Feas.\,(\%)} & \textbf{R.\,fail\,(\%)} & \textbf{Med.\ gap\,(\%)} \\
\midrule
\textsc{SDDL}   & JSSP     & 100 & 96.0 & 3.0  & 0.0 \\
\textsc{SDDL}   & SM-RCPSP & 100 & 17.0 & 44.0 & 0.0 \\
\textsc{SDDL}   & MM-RCPSP & 100 & 53.0 & 1.0  & 0.0 \\
\textsc{Solver} & JSSP     & 100 & 53.0 & 40.0 & 5.4 \\
\textsc{Solver} & SM-RCPSP & 100 & 7.0  & 68.0 & 0.0 \\
\textsc{Solver} & MM-RCPSP & 100 & 5.0  & 80.0 & 0.0 \\
\textsc{Direct} & JSSP     & 100 & 52.0 & 0.0  & 697.4 \\
\textsc{Direct} & SM-RCPSP & 100 & 4.0  & 0.0  & 0.0 \\
\textsc{Direct} & MM-RCPSP & 100 & 15.0 & 0.0  & 0.0 \\
\bottomrule
\end{tabular}
\caption{qwen3.5-27b by family and condition ($N=300$ convention; run-fail
includes instances without a completed generation).}
\label{tab:app-family-qwen}
\end{table*}

\begin{table*}[p]
\centering\footnotesize
\setlength{\tabcolsep}{5pt}
\begin{tabular}{@{}lrrrrr@{}}
\toprule
\textbf{Failure code} & \textbf{qwen3.5} & \textbf{devstral} & \textbf{q3-coder} & \textbf{magistral} & \textbf{Total} \\
\midrule
Dropped / garbled precedence edges (tasks complete)      & 6 & 4 & 4 & 5 & 19 \\
Referenced-but-undeclared real activity                  & 1 & 2 & 0 & 3 & 6 \\
Phantom identifier (no problem counterpart)              & 1 & 1 & 0 & 0 & 2 \\
Duplicate task declarations                              & 0 & 1 & 1 & 0 & 2 \\
Coverage collapse / label mismatch                       & 0 & 1 & 2 & 0 & 3 \\
Primitive-argument misuse                                & 0 & 0 & 1 & 0 & 1 \\
Emission derailment (prose drift, repetition, syntax slip) & 2 & 0 & 0 & 1 & 3 \\
\midrule
Coded / sampled & 10/10 & 9/10 & 8/10 & 9/10 & 36/40 \\
\bottomrule
\end{tabular}
\caption{Hand-coded causes over a stratified sample of \textsc{SDDL}
failures (one primary code per instance).}
\label{tab:app-taxonomy}
\end{table*}

\clearpage
\onecolumn
\noindent\textbf{The \textsc{SDDL} system prompt.}\label{lst:app-prompt-sddl}
\begin{lstlisting}[basicstyle=\ttfamily\scriptsize,breaklines=true,breakindent=0pt,columns=fullflexible,frame=single,framesep=3pt,xleftmargin=2pt,xrightmargin=2pt]
You are completing an automated benchmark. The user message contains a scheduling problem written in natural language, followed by a "Response Format" section describing schedule lines. IGNORE that Response Format section entirely -- it describes a different answer mode and does not apply to you. Do not solve the problem and do not emit a schedule. Instead, translate the problem into SDDL: a small declarative language that a downstream compiler turns into an exact solver model. The first character of your response must be the first character of the DSL; output nothing else -- no preamble, no explanation, no reasoning, no markdown, no code fences, no XML tags, no comments.

Every piece of instance data must be transcribed from the user message into the DSL as literals. Transcription accuracy is the whole task: a single wrong duration or machine silently produces a valid-looking but wrong answer. Do not omit any task, any precedence, or any resource.

## The 7 primitives

resource(id, **props)      a machine, renewable resource, or nonrenewable budget
task(id, **props)          one operation / activity to schedule
before(a, b)               HARD: task a finishes before task b starts
no_overlap(resource_id)    HARD: one task at a time on that resource
conflict(t1, ..., group=)  HARD: listed tasks cannot share a time slot
not_at(id, day=, period=)  HARD: the task cannot run at that time
penalize(measure, weight=1)      soft objective; measures: makespan,
                           capacity, spread, isolated, room_stability

## Task properties

- duration=N          time length
- label="Exact Name"  REQUIRED on every task. See the Labels section below.
- machine="rid"       fixed resource assignment; pair with no_overlap(rid)
- job=N, position=N   job index and 0-BASED operation index within that job
- demands={"r": n}    renewable units used WHILE active
- consumes={"b": n}   nonrenewable units used ONCE
- modes=[{"duration": N, "demands": {...}, "consumes": {...}}, ...]
                      alternative execution modes; the solver picks exactly one

## Resource properties

- capacity=N   renewable: per-time-period limit
- total=N      nonrenewable: project-wide budget

## Labels -- read carefully, results are discarded without them

`label` carries the entity's display name from the problem text, verbatim:
capitalization, spacing, and punctuation exactly as written.

- Job-shop problems: label is the ITEM / JOB name ONLY, never with a step
  suffix; the 0-based step index goes in position=.
- Project problems: label is the ACTIVITY name, including the named start and
  finish milestones. Every activity gets its own task() with its own label.

## Identifiers

`id` and resource ids are sanitized snake_case, never the display name. Keys of
demands={} / consumes={} MUST exactly match the corresponding resource() id.

## Rules

1. One task() per operation (job shop) or per activity (project scheduling),
   including start and finish milestones.
2. One resource() per machine, renewable resource, or nonrenewable budget.
3. One before() per SUCCESSOR ENTRY. Work through the activities in order; for
   each one, transcribe all of its listed successors before moving to the next.
   An activity listing k successors contributes k before() calls -- do not move
   on with a partial successor list. Job-shop steps are chained consecutively.
4. One no_overlap() per job-shop machine. NEVER use no_overlap for a project
   scheduling renewable resource -- capacity= plus demands={} already gives the
   correct cumulative semantics, and adding it makes the model wrong.
5. A mode dict omits resources it does not use. An activity with one fixed mode
   uses plain duration=/demands=/consumes=, not modes=[].
6. End with one penalize() per stated objective, naming the measure that
   matches the problem's objective and its weight.
7. Output ONLY the DSL calls.
\end{lstlisting}

\appendix

\end{document}